%% file: main.tex
\documentclass{article}

\usepackage{spconf}
\input{preamble}

\title{CASE: Complementary Articulation-aware Speech Encoder to Enhance Speech-Driven 3D Facial Animation}

\name{
Jaehong Myung$^{1}$,
Seungho Eum$^{2}$,
Taegon Kim$^{2}$,
Hyunjung Chung$^{2}$,
and Unsang Park$^{1,2\dagger}$
\thanks{$^{\dagger}$Corresponding author: Unsang Park (unsangpark@sogang.ac.kr).}
}
\address{
$^{1}$Department of Artificial Intelligence, Sogang University, Seoul, Republic of Korea\\
$^{2}$Department of Computer Science and Engineering, Sogang University, Seoul, Republic of Korea
}

\begin{document}
\ninept
\maketitle

\input{sec/00_abstract}
\input{sec/01_intro}
\input{sec/02_method}
\input{sec/03_experiments}
\input{sec/04_limitations}
\input{sec/05_conclusion}

\clearpage
\icasppolicyheading{Acknowledgments}
This work was supported by the Institute of Information \& Communications
Technology Planning \& Evaluation (IITP) grant funded by the Korea
government (MSIT) (No. 2022-0-00621, RS-2022-II220621,
Development of artificial intelligence technology that provides
dialog-based multi-modal explainability).

\noindent\textbf{Conflict of Interest.}
The authors have no relevant financial or non-financial conflicts of interest to disclose.

\icasppolicyheading{Compliance with Ethical Standards}
This study uses publicly released datasets and involves no new collection
of human-participant or animal data.

\bibliographystyle{IEEEbib}
\bibliography{strings,main}

\end{document}

%% file: preamble.tex
\usepackage{amsmath,amssymb}
\usepackage{graphicx}
\usepackage{booktabs}
\usepackage{multirow}
\usepackage{makecell}
\usepackage{balance}
\usepackage{placeins}
\usepackage{float}
\usepackage{cuted}
\usepackage{capt-of}
\usepackage[hidelinks]{hyperref}

\newcommand{\CASE}{\textsc{Case}}
\newcommand{\AVSE}{\textsc{AV-SE}}
\newcommand{\Hmain}{\mathbf{H}_{\mathrm{main}}}
\newcommand{\Hcase}{\widetilde{\mathbf{H}}_{\mathrm{CASE}}}

\newcommand{\icasppolicyheading}[1]{%
  \par\medskip
  {\centering\bfseries\MakeUppercase{#1}\par}%
  \smallskip
}

%% file: sec/00_abstract.tex
\begin{abstract}
General-purpose self-supervised speech encoders provide strong acoustic
representations, but their pretraining objectives are not explicitly
aligned with the visible articulatory dynamics required by speech-driven
3D facial animation. We propose \CASE{} (Complementary
Articulation-aware Speech Encoder), a complementary speech encoder
adapted by contrastively aligning short speech clips with continuous 3D
lip-mesh VQ-VAE latents. After adaptation, the geometry branch is
removed, and the audio-only \CASE{} representation is fused
with the baseline SSL speech representation through residual
cross-attention, while retaining the motion-decoder architecture and objective. Across four VOCASET baselines, \CASE{} reduces mean lip vertex error by 2.8--6.0\%, and mean full-face vertex error decreases in every system. Under the same fusion protocol, \CASE{} yields lower mean LVE than the unadapted \AVSE{} stream on all four generators, with model-dependent margins. Because adaptation uses Korean speech--geometry pairs and evaluation uses English VOCASET, the result is consistent with cross-language downstream transfer under a shared English-pretrained audio-visual prior.
\end{abstract}


%% file: sec/01_intro.tex
\section{Introduction}
\label{sec:intro}

Speech-driven 3D facial animation maps speech to temporally aligned 3D
face meshes~\cite{cudeiro2019capture,fan2022faceformer,
xing2023codetalker}. The pre-trained SSL speech encoder determines which
acoustic and linguistic cues are exposed to the motion generator.
FaceFormer and CodeTalker use wav2vec~2.0~\cite{fan2022faceformer,
xing2023codetalker,baevski2020wav2vec}, SelfTalk uses the multilingual
XLSR-53 representation~\cite{peng2023selftalk,conneau2021xlsr}, and
UniTalker adopts WavLM~\cite{fan2024unitalker,chen2022wavlm}. These
systems span different self-supervised speech-pretraining regimes, yet
their generators are still conditioned on a single baseline SSL speech
representation.

Self-supervised speech representations can contain linearly recoverable
articulatory kinematics~\cite{cho2024self}. Audio-visual
self-supervised models further exploit synchronized speech and 2D lip
video to shape speech representations~\cite{shi2022learning}, but these
objectives do not explicitly target local continuous 3D lip geometry.
Acoustic and visible distinctions are also not identical: /p/ and /b/
differ through voicing while sharing bilabial closure, and vowel mouth
shapes vary with coarticulatory context~\cite{birkholz2013modeling}.
This leads to the central question of this work: can a complementary
speech encoder adapted to local 3D lip geometry using Korean
speech--mesh pairs provide useful features for English facial animation
without redesigning the motion-decoder architecture or objective?

Neighboring approaches introduce articulation-related information
through different routes. FaceXHuBERT directly conditions expressive
3D facial animation on a self-supervised speech representation
~\cite{haque2023facexhubert}. PhISANet adds phone-level CTC supervision
to an end-to-end animation model~\cite{medina2024phisanet}, whereas
RQTalker uses region-aware vector quantization as a generator-side
motion prior~\cite{fan2025rqtalker}. Wav2Sem supplies a reusable
semantic audio branch~\cite{li2025wav2sem}; other methods use a
lip-reading expert as a perceptual loss~\cite{eungi2024enhancing} or
learn a speech--mesh representation for perceptual supervision and
evaluation~\cite{chae2025perceptually}. In contrast, \CASE{} requires
neither phone labels nor a replacement motion-generation architecture.
It adapts a complementary speech encoder to local 3D lip latents and
injects its audio-only features before an existing generator.

We learn \CASE{} by clip-level contrastive alignment between speech and
continuous 3D lip-mesh VQ-VAE~\cite{van2017neural} latents. The geometry
branch is discarded after training, leaving a reusable \CASE{} encoder
whose local features are fused with the baseline SSL speech
representation. Our contributions are: (1) \CASE{}, a complementary
speech encoder that adapts a 2D audio-visual speech prior to local 3D lip
geometry through clip-level contrastive learning; (2) a
decoder-preserving residual cross-attention interface that injects
audio-only \CASE{} features into heterogeneous facial-animation
systems; and (3) a controlled four-generator comparison with the
unadapted \AVSE{} stream and online profiling, evaluating transfer from
Korean 3D adaptation to English VOCASET and deployment cost.

%% file: sec/02_method.tex
\section{Method}
\label{sec:method}

\begin{figure*}[t]
    \centering
    \includegraphics[width=0.955\textwidth]{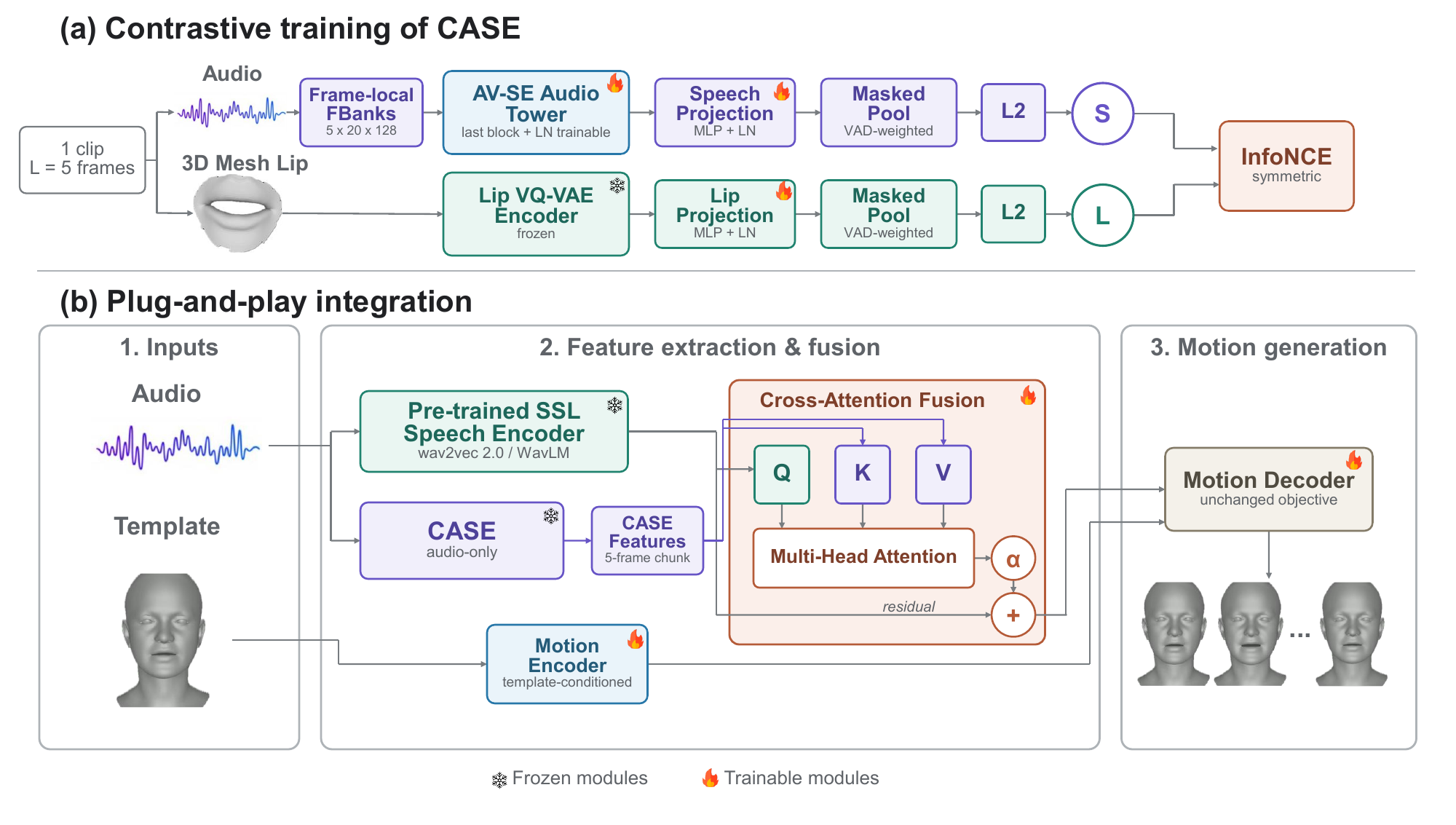}
    \caption{Overview of \CASE{}. (a) Clip-level contrastive training
    aligns frame-local speech features with continuous 3D lip-mesh
    latents. (b) After adaptation, the geometry branch is removed and
    features from the audio-only \CASE{} encoder are fused through
    residual cross-attention before the original motion-generation
    stack. The baseline block denotes a pre-trained SSL speech
    encoder---wav2vec~2.0, XLSR-53, or WavLM in our
    experiments~\cite{baevski2020wav2vec,conneau2021xlsr,chen2022wavlm}.
    The original motion-generation architecture and training objective
    are retained. Snowflakes and flames denote frozen and trainable
    modules.}
    \label{fig:overview}
\end{figure*}

\subsection{Articulation-Aware Speech Adaptation}
\label{sec:adaptation}

As shown in Fig.~\ref{fig:overview}(a), \CASE{} adapts the speech
branch of \AVSE{} using local 3D lip geometry as cross-modal
supervision. \AVSE{} is the stage-1 audio-visual speech encoder
of~\cite{chae2025perceptually}, learned from synchronized speech and 2D
lip video. We use the publicly released Training partition of the Korean
\emph{3D Audio-visual Talking Face Dataset}, distributed through
AI-Hub, South Korea, comprising ten speakers under an 8/1/1
speaker-disjoint split for training, validation, and testing
(M01--M08/M09/M10). From each 30-fps mesh frame, we extract a
4,410-vertex mouth region. For a lip mesh $\mathbf{y}_t$,
\begin{equation}
\mathbf{z}_t=E_{\mathrm{lip}}(\mathbf{y}_t)
\label{eq:lip_latent}
\end{equation}
is the 1024-d continuous pre-quantization output of a separately
trained and frozen lip-centric VQ-VAE. Thus the target is local
geometry rather than a discrete code or lexical label.

Each five-frame mesh clip ($\sim$167~ms) is paired with five
frame-local $20\times128$ filterbanks, so adaptation is driven by temporally local
speech--articulation correspondence rather than an utterance-level
semantic target. Lightweight projection heads $g_s$ and $g_l$ map the
\AVSE{} speech features $\phi(\mathbf{x}_{i,t})$ and lip latents
$\mathbf{z}_{i,t}$ into a shared 512-dimensional space; VAD-masked
pooling then produces clip-level embeddings:
\begin{equation}
\begin{aligned}
\bar{\mathbf{s}}_i &= \mathrm{Norm}\!\left(
\frac{\sum_t m_{i,t}g_s(\phi(\mathbf{x}_{i,t}))}
{\sum_t m_{i,t}}\right),\\
\bar{\mathbf{l}}_i &= \mathrm{Norm}\!\left(
\frac{\sum_t m_{i,t}g_l(\mathbf{z}_{i,t})}
{\sum_t m_{i,t}}\right).
\end{aligned}
\label{eq:pool}
\end{equation}
Five-frame clips are centered on speech-active regions and retained
only when at least three frames are voiced. For a mini-batch of $B$
matched clips, we optimize symmetric InfoNCE
~\cite{oord2018representation}:
\begin{equation}
\begin{aligned}
\mathcal{L}_{\mathrm{con}}
&=\tfrac{1}{2}(\mathcal{L}_{s\rightarrow l}+\mathcal{L}_{l\rightarrow s}),\\
\mathcal{L}_{s\rightarrow l}
&=-\frac{1}{B}\sum_i\log
\frac{e^{\mathrm{sim}(\bar{\mathbf{s}}_i,\bar{\mathbf{l}}_i)/\tau}}
{\sum_j e^{\mathrm{sim}(\bar{\mathbf{s}}_i,\bar{\mathbf{l}}_j)/\tau}}.
\end{aligned}
\label{eq:infonce}
\end{equation}
Here $\tau$ is the temperature; the reverse direction is defined
analogously. All other clips in the mini-batch serve as negatives in
both directions. We train the adaptation for 12 epochs with AdamW,
cosine decay, and 5\% warm-up. Only the final \AVSE{} Transformer
block, its output normalization, and the two projection heads are
updated; the remaining speech backbone and VQ-VAE are frozen. This
restricted update preserves the audio-visual prior while adapting a
small parameter subset to the 3D corpus.

\subsection{Plug-and-Play Integration}

After training, \CASE{} operates as an audio-only complementary speech
encoder: the lip branch, VQ-VAE, and VAD are removed from the deployed
path. \CASE{} produces one 512-dimensional feature for each
non-overlapping five-frame output chunk and aligns the resulting
sequence to the baseline timeline.

Let $\Hmain$ be the output of the baseline SSL speech encoder and
$\Hcase$ the aligned \CASE{} feature stream. Residual cross-attention is
\begin{equation}
\begin{aligned}
\mathbf{Q}&=W_Q\mathrm{LN}_m(\Hmain),\\
\mathbf{K}&=W_K\mathrm{LN}_c(\Hcase),\quad
\mathbf{V}=W_V\mathrm{LN}_c(\Hcase),\\
\mathbf{H}_{\mathrm{fused}}
&=\Hmain+\alpha W_O\operatorname{MHA}(\mathbf{Q},\mathbf{K},\mathbf{V}).
\end{aligned}
\label{eq:fusion}
\end{equation}
The baseline SSL speech representation provides the queries, while
\CASE{} features provide the keys and values. The four-head module uses
independent stream normalizations; $\alpha$ is initialized to 0.1 and
$W_O$ to zero, so integration starts as an exact identity. Fusion is
applied to the output of the baseline SSL speech encoder; the
baseline-specific projection and motion-generation stack retain their
original architecture and training objective. The template-conditioned
path and generation procedure are also retained.

%% file: sec/03_experiments.tex
\begin{table*}[t]
\centering
\renewcommand{\arraystretch}{1.06}
\setlength{\tabcolsep}{10.5pt}
\caption{Main VOCASET results (mean over three runs). Bold denotes
the lower value within each baseline pair. Lip Vertex Error is the
primary articulation-sensitive metric; Mean Vertex Error measures
full-face reconstruction.}
\label{tab:main_results}
\begin{tabular}{@{}llccc@{}}
\toprule
Baseline (SSL Encoder) & Method &
\makecell{Lip Vertex Error\\($\times10^{-5}$) $\downarrow$} &
\makecell{Mean Vertex Error\\($\times10^{-6}$) $\downarrow$} &
\makecell{Relative LVE\\Change} \\
\midrule
\multirow{2}{*}{\makecell[l]{CodeTalker~\cite{xing2023codetalker}\\wav2vec~2.0~\cite{baevski2020wav2vec}}}
& Baseline & 3.8744 & 2.1171 & -- \\
& +\CASE{} & \textbf{3.6424} & \textbf{1.9764} & $-6.0\%$ \\
\midrule
\multirow{2}{*}{\makecell[l]{FaceFormer~\cite{fan2022faceformer}\\wav2vec~2.0~\cite{baevski2020wav2vec}}}
& Baseline & 3.4864 & 1.9738 & -- \\
& +\CASE{} & \textbf{3.3677} & \textbf{1.9236} & $-3.4\%$ \\
\midrule
\multirow{2}{*}{\makecell[l]{SelfTalk~\cite{peng2023selftalk}\\XLSR-53~\cite{conneau2021xlsr}}}
& Baseline & 2.9629 & 1.7719 & -- \\
& +\CASE{} & \textbf{2.8213} & \textbf{1.7653} & $-4.8\%$ \\
\midrule
\multirow{2}{*}{\makecell[l]{UniTalker~\cite{fan2024unitalker}\\WavLM~\cite{chen2022wavlm}}}
& Baseline & 3.3778 & 1.9342 & -- \\
& +\CASE{} & \textbf{3.2832} & \textbf{1.8875} & $-2.8\%$ \\
\bottomrule
\end{tabular}
\end{table*}

\section{Experiments}
\label{sec:experiments}

\subsection{Setup and Metrics}

We evaluate all four systems on a common VOCASET subject
split~\cite{cudeiro2019capture}. The systems expose three baseline SSL
speech representations: wav2vec~2.0, XLSR-53, and WavLM. For
UniTalker, we use a VOCASET-only instantiation with a single output
head, retain its 512-dimensional PCA motion representation, and
initialize its audio encoder from WavLM-base-plus
~\cite{fan2024unitalker}. Within each evaluated system, the Baseline,
+\AVSE{}, and +\CASE{} conditions retain the same generator
architecture, training objective, and downstream schedule; the two
auxiliary conditions additionally share the same fusion interface. The
\CASE{} encoder in all reported experiments is the 12-epoch checkpoint
described in Sec.~\ref{sec:adaptation}.

For aligned prediction and ground-truth sequences of length $T$, let
$\mathbf{v}_{t,i}$ and $\widehat{\mathbf{v}}_{t,i}$ denote the
corresponding 3D vertex positions. Following
MeshTalk~\cite{richard2021meshtalk}, we report lip vertex error (LVE)
as the primary articulation-sensitive metric. For each frame, LVE uses
the maximum squared $\ell_2$ error over the lip vertices, since spatial
averaging can mask locally inaccurate lip shapes. We additionally
report mean vertex error (MVE) as a complementary measure of full-face
reconstruction:
\begin{equation}
\begin{aligned}
\mathrm{LVE} &= \frac{1}{T}\sum_{t=1}^{T}
\max_{i\in\mathcal{V}_{\mathrm{lip}}}
\|\mathbf{v}_{t,i}-\widehat{\mathbf{v}}_{t,i}\|_2^2,\\
\mathrm{MVE} &= \frac{1}{T}\sum_{t=1}^{T}\frac{1}{N}
\sum_{i=1}^{N}
\|\mathbf{v}_{t,i}-\widehat{\mathbf{v}}_{t,i}\|_2^2.
\end{aligned}
\label{eq:metrics}
\end{equation}
Here $\mathcal{V}_{\mathrm{lip}}$ is the lip-vertex set and $N$ is the
total number of facial vertices; lower is better. On held-out Korean
data, the lip VQ-VAE reconstructs the target region with per-vertex MSE
$1.14\times10^{-6}$ and uses 215 of 256 codes (perplexity 107.9).

\subsection{Main Results}

Table~\ref{tab:main_results} shows lower mean LVE for all four
generators. The relative reductions are 6.0\% for CodeTalker, 3.4\%
for FaceFormer, 4.8\% for SelfTalk, and 2.8\% for UniTalker. Mean MVE
also decreases by 6.6\%, 2.5\%, 0.4\%, and 2.4\%, respectively.
Because \CASE{} is supervised by lip-local geometry, the LVE reductions
are the primary result; the accompanying MVE reductions indicate that
these local gains do not come at the expense of full-face
reconstruction.

The systems expose wav2vec~2.0, XLSR-53, and WavLM and span
discrete-prior, autoregressive, and non-autoregressive generators. The
consistent mean LVE reduction indicates that the 3D-adapted \CASE{}
stream remains useful on English VOCASET across heterogeneous speech
representations and generators, with model-dependent effect sizes.

\begin{table}[t]
\centering
\renewcommand{\arraystretch}{1.04}
\caption{3D-adaptation control. +\AVSE{}~\cite{chae2025perceptually}
uses the English-pretrained AV-SE encoder; +\CASE{} uses the same
initialization, fusion, and insertion point after 12-epoch adaptation to
Korean 3D lip latents.}
\label{tab:avse_ablation}
\begin{tabular*}{\columnwidth}{@{\extracolsep{\fill}}llcc@{}}
\toprule
Baseline & Method &
\makecell{Lip Vertex Error\\($\times10^{-5}$) $\downarrow$} &
$\Delta$LVE \\
\midrule
\multirow{3}{*}{CodeTalker~\cite{xing2023codetalker}}
& Baseline & 3.8744 & -- \\
& +\AVSE{} & 3.7535 & $-3.1\%$ \\
& +\CASE{} (ours) & \textbf{3.6424} & \textbf{$-6.0\%$} \\
\midrule
\multirow{3}{*}{FaceFormer~\cite{fan2022faceformer}}
& Baseline & 3.4864 & -- \\
& +\AVSE{} & 3.3858 & $-2.9\%$ \\
& +\CASE{} (ours) & \textbf{3.3677} & \textbf{$-3.4\%$} \\
\midrule
\multirow{3}{*}{SelfTalk~\cite{peng2023selftalk}}
& Baseline & 2.9629 & -- \\
& +\AVSE{} & 2.9379 & $-0.8\%$ \\
& +\CASE{} (ours) & \textbf{2.8213} & \textbf{$-4.8\%$} \\
\midrule
\multirow{3}{*}{UniTalker~\cite{fan2024unitalker}}
& Baseline & 3.3778 & -- \\
& +\AVSE{} & 3.2947 & $-2.5\%$ \\
& +\CASE{} (ours) & \textbf{3.2832} & \textbf{$-2.8\%$} \\
\bottomrule
\end{tabular*}
\end{table}

\subsection{3D Adaptation under a Shared AV-SE Prior}

\AVSE{} is trained on synchronized English speech and 2D lip video,
providing a strong audio-visual prior. Table~\ref{tab:avse_ablation}
compares it with the same stream after 3D lip-geometry adaptation using
Korean data. Relative to Baseline, +\AVSE{} lowers mean LVE by 3.1\%,
2.9\%, 0.8\%, and 2.5\%; \CASE{} yields further mean reductions of
3.0\%, 0.5\%, 4.0\%, and 0.3\%. At the level of reported means,
\CASE{} is lower than +\AVSE{} on all four systems, although the
difference varies by model.

Because the two conditions share initialization and integration, their
mean gap provides a controlled downstream comparison of 3D lip-geometry
adaptation.

\begin{figure*}[!t]
\centering
\includegraphics[width=0.985\textwidth,trim=24 43 30 31,clip]{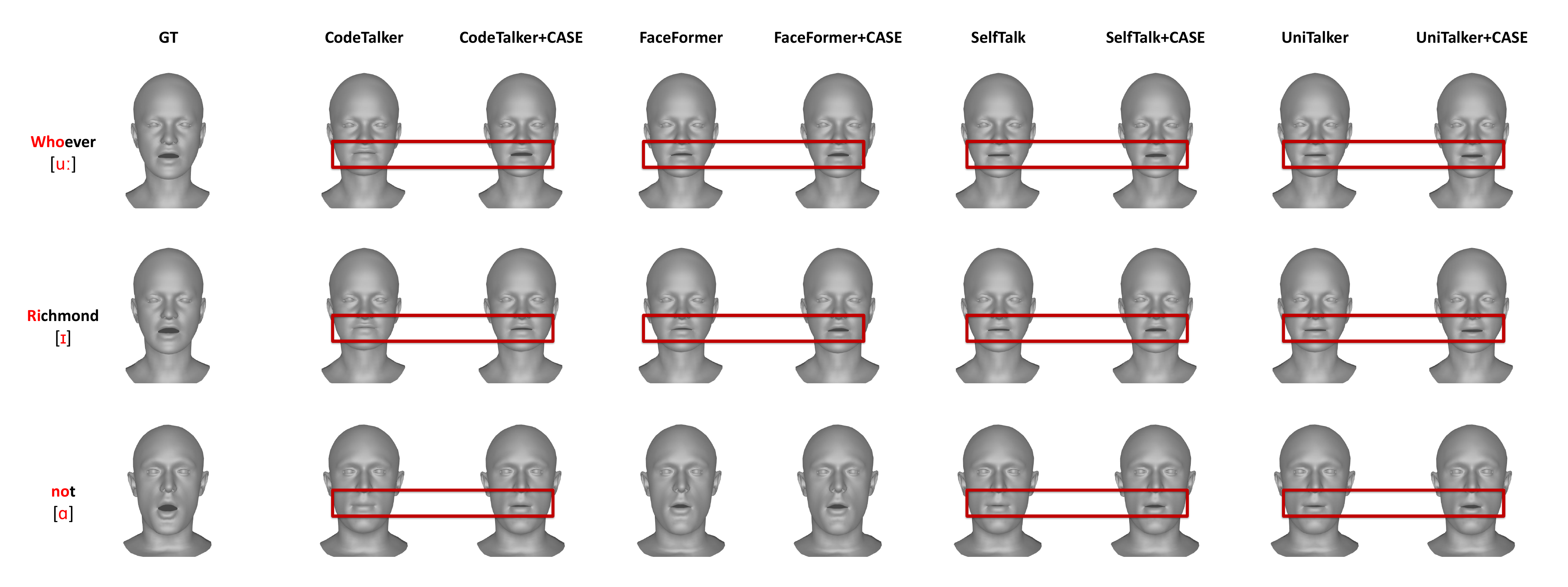}
\caption{Illustrative qualitative comparisons on selected VOCASET
frames. Each row shows the ground truth and paired Baseline/+\CASE{}
outputs for CodeTalker, FaceFormer, SelfTalk, and UniTalker. Red boxes
mark the mouth regions selected for visual comparison.}
\label{fig:qualitative}
\vspace{3pt}
\centering
\captionof{table}{End-to-end inference on 39 VOCASET test utterances
(RTX~3090). Latency is the mean of per-utterance medians over three
post-warm-up passes; $\Delta$VRAM is relative to the original baseline.
Added latency is +\CASE{} minus original inference; runtime/audio below
1 means processing is shorter than the input audio.}
\label{tab:runtime}
\renewcommand{\arraystretch}{0.88}
\setlength{\tabcolsep}{5.0pt}
\begin{tabular}{@{}lrrrrrr@{}}
\toprule
Generator & \makecell{Original\\latency (ms)} &
\makecell{+\CASE{}\\latency (ms)} &
\makecell{Added \CASE{}\\latency (ms)} &
\makecell{Runtime / audio\\duration ($\times$)} &
\makecell{$\Delta$VRAM\\(MB)} &
\makecell{Fusion $\Delta$ params\\(\% trainable)} \\
\midrule
CodeTalker~\cite{xing2023codetalker} & 1650 & 1799 & +149 & 0.47 & +132 & 1.97M (1.08\%) \\
FaceFormer~\cite{fan2022faceformer} & 230 & 405 & +175 & 0.10 & +132 & 1.97M (2.14\%) \\
SelfTalk~\cite{peng2023selftalk} & 21 & 199 & +178 & 0.05 & +135 & 3.15M (0.70\%) \\
UniTalker~\cite{fan2024unitalker} & 22 & 716 & +694 & 0.19 & +133 & 1.97M (2.16\%) \\
\bottomrule
\end{tabular}
\end{figure*}

\subsection{Qualitative Comparison}

Figure~\ref{fig:qualitative} shows illustrative local mouth-shape changes
for the four evaluated systems. The paired outputs visualize differences
in lip aperture and rounding.

\subsection{End-to-End Online Inference Cost}

Table~\ref{tab:runtime} reports the full online path for unseen audio,
including \CASE{} feature extraction. \CASE{} adds 149--178~ms to
CodeTalker, FaceFormer, and SelfTalk. The current UniTalker
implementation adds 694~ms because it recomputes filterbanks for each
five-frame chunk on the CPU. Nevertheless, total processing time
remains below the input-audio duration for all four systems, with
runtime-to-audio ratios between 0.05 and 0.47. Additional memory is
limited to 132--135~MB, and the fusion module increases trainable
parameters by only 0.70--2.16\%.

%% file: sec/04_limitations.tex
\section{Limitations and Future Work}
\label{sec:limitations}

Our evaluation is currently limited to VOCASET. BIWI, a commonly used
alternative benchmark for speech-driven 3D facial animation
~\cite{fanelli2010biwi}, was not publicly available to us during this
study, preventing a matched second-dataset evaluation. Evaluating
\CASE{} on additional datasets and establishing a Korean downstream
benchmark would provide broader tests of cross-dataset and
cross-language generalization.

\CASE{} starts from an English-pretrained \AVSE{} encoder. \CASE{}
has lower mean LVE than +\AVSE{} on all four generators, but the
difference varies by model. Because the 2D prior and 3D adaptation remain
intertwined, this Korean-to-English result is consistent with
cross-language transfer under a shared prior, not language-independent
articulation. Future work can examine multilingual or non-English priors
and other SSL speech encoders.

The current evidence for articulation awareness is primarily
downstream, through lower mean LVE across the four systems and
illustrative local mouth-shape changes. Direct analyses of lip closure,
rounding, and coarticulatory trajectories would provide a stronger
mechanistic account of which articulatory properties are encoded by
\CASE{}.

%% file: sec/05_conclusion.tex
\section{Conclusion}
\label{sec:conclusion}

\CASE{} adapts an English-pretrained audio-visual speech prior to local
3D lip geometry and injects the resulting audio-only features into
existing generators without changing their decoders or objectives.
Across four generators, it reduces mean LVE by 2.8--6.0\% and lowers
mean MVE in every system. It also yields lower mean LVE than unadapted
\AVSE{}, with model-dependent margins. Adaptation uses Korean
speech--geometry pairs and evaluation uses English VOCASET, making the
result consistent with cross-language transfer under a shared prior.
Online processing remains shorter than the input audio, supporting
practical reuse in existing systems.